\documentclass[sigconf,screen=true]{acmart}

\usepackage{balance}
\usepackage{subcaption}
\usepackage{colortbl}        
\usepackage[most]{tcolorbox}
\usepackage{multirow}
\usepackage{makecell}
\usepackage{tabularx}
\definecolor{ForestGreen}{rgb}{0, 0.69, 0.31}
\definecolor{NavyBlue}{rgb}{0, 0.44, 0.75}
\newcommand{\hgreen}[1]{\textcolor{ForestGreen}{\textbf{#1}}} 
\definecolor{lightblue}{rgb}{0.5,0.6,0.7}  

\usepackage{tikz}
\usepackage{enumitem}
\newcommand{\ie}{\emph{i.e., }}
\newcommand{\eg}{\emph{e.g., }}
\newcommand{\etal}{\emph{et al.}}

\AtBeginDocument{%
  }

\copyrightyear{2026}
\acmYear{2026}
\setcopyright{cc}
\setcctype{by-nc-nd}
\acmConference[KDD '26]{Proceedings of the 32nd ACM SIGKDD Conference on Knowledge Discovery and Data Mining V.1}{August 09--13, 2026}{Jeju Island, Republic of Korea}
\acmBooktitle{Proceedings of the 32nd ACM SIGKDD Conference on Knowledge Discovery and Data Mining V.1 (KDD '26), August 09--13, 2026, Jeju Island, Republic of Korea}
\acmPrice{}
\acmDOI{10.1145/3770854.3783934}
\acmISBN{979-8-4007-2258-5/2026/08}

\begin{document}

\title{Towards Trustworthy Multimodal Moderation via Policy-Aligned Reasoning and Hierarchical Labeling}

\author{Anqi Li}
\email{anqi.li@sjtu.edu.cn}
\affiliation{%
  \institution{Shanghai Jiao Tong University}
  \city{Shanghai}
  \country{China}
}

\author{Wenwei Jin}
\email{wenwei1217.jin@gmail.com}
\authornote{Project lead.}
\affiliation{%
  \institution{Xiaohongshu Inc.}
  \city{Beijing}
  \country{China}
}

\author{Jintao Tong}
\email{jintaotong@hust.edu.cn}
\affiliation{%
  \institution{Huazhong University of Science and Technology}
  \city{Wuhan}
  \country{China}
}

\author{Pengda Qin}
\email{qinpengda0406@gmail.com}
\affiliation{%
  \institution{Xiaohongshu Inc.}
  \city{Beijing}
  \country{China}
}

\author{Jiawei Li}
\email{wangdesheng@xiaohongshu.com}
\affiliation{%
  \institution{Xiaohongshu Inc.}
  \city{Hangzhou}
  \country{China}
}

\author{Guo Lu}
\email{luguo2014@sjtu.edu.cn}
\authornote{Corresponding author.}
\affiliation{%
  \institution{Shanghai Jiao Tong University}
  \city{Shanghai}
  \country{China}
  }

\renewcommand{\shortauthors}{Anqi Li et al.}


\begin{abstract}
Social platforms have revolutionized information sharing, but also accelerated the dissemination of harmful and policy-violating content. To ensure safety and compliance at scale, moderation systems must go beyond efficiency and offer accuracy and interpretability. However, current approaches largely rely on noisy, label-driven learning, lacking alignment with moderation rules and producing opaque decisions that hinder human review.~
Therefore, we propose \textbf{Hi}erarchical \textbf{Guard} (\textbf{Hi-Guard}), a multimodal moderation framework that introduces a new policy-aligned decision paradigm. 
The term “Hierarchical” reflects two key aspects of our system design: (1) a hierarchical moderation pipeline, where a lightweight binary model first filters safe content and a stronger model handles fine-grained risk classification; and (2) a hierarchical taxonomy in the second stage, where the model performs path-based classification over a hierarchical taxonomy ranging from coarse to fine-grained levels.~
To ensure alignment with evolving moderation policies, Hi-Guard directly incorporates rule definitions into the model prompt. To further enhance structured prediction and reasoning, we introduce a multi-level soft-margin reward and optimize with Group Relative Policy Optimization (GRPO), penalizing semantically adjacent misclassifications and improving explanation quality.~Extensive experiments and real-world deployment demonstrate that Hi-Guard achieves superior classification accuracy, generalization, and interpretability, paving the way toward scalable, transparent, and trustworthy content safety systems. Code is available at: \href{https://github.com/lianqi1008/Hi-Guard}{https://github.com/lianqi1008/Hi-Guard}.

\end{abstract}

\ccsdesc[500]{Security and privacy~Social network security and privacy}
\ccsdesc[300]{Computing methodologies~Computer vision}
\ccsdesc[300]{Computing methodologies~Natural language processing}
\ccsdesc[100]{Computing methodologies~Reinforcement learning}

\keywords{Multimodal LLMs, Hierarchical Classification, Multimodal Content Safety, Reinforcement Learning, Explainable AI}
  

\maketitle

\begin{figure*}[t]
  \includegraphics[width=\linewidth]{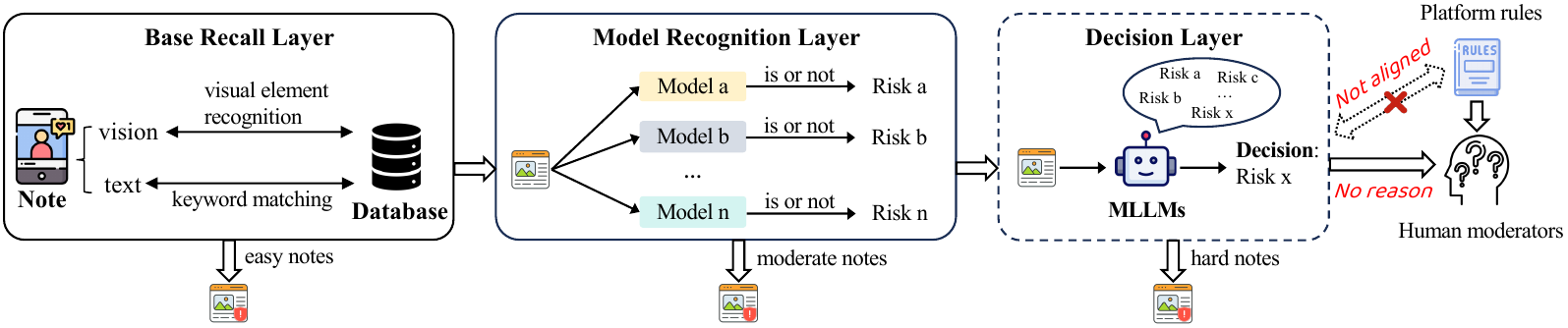}
  \caption{Moderation system architecture. The term “layer” here refers to functional stages in the moderation pipeline. Stages enclosed by solid lines are inherent components of the system, while dashed boxes indicate the stage we aim to optimize.}
  \label{fig:system}
\end{figure*}

\section{Introduction}\label{sec:intro}
The ever-expanding scale and complexity of user-generated content (UGC) on social media platforms have created immense challenges for content moderation~\cite{gerrard2020behind, ruckenstein2020re}. With millions of posts shared daily, ranging from text and images to video~\cite{wang2024multihateclip}, platforms must continuously evolve their moderation system~\cite{ma2021advertiser} to detect increasingly subtle harmful content~\cite{maisto2017mining,livingstone2014their,citron2014hate,fernandez2018online} and emerging risk types while maintaining a safe environment and user experience.~
To meet this demand, large-scale platforms have adopted multi-stage moderation pipelines comprising a base recall layer, a model-based recognition layer, and a final human review. With the superior comprehension capabilities and impressive performance across a wide range of multimodal content understanding tasks, multimodal large language models (MLLMs)~\cite{liu2024improved,dai2023instructblip,achiam2023gpt,bai2025qwen2, zhang2025videollama, team2024gemini, anthropic2024claude} have been widely utilized in real-world content moderation scenarios~\cite{wu2025icm, hee2024recent,jo2024harmful,jin2024mmsoc}. As illustrated in Figure~\ref{fig:system}, MLLMs are deployed in the decision layer to analyze difficult UGC cases (referred to as \emph{Note} on our platform) and determine both the presence and type of risk, playing a pivotal role in the final moderation decision.

Despite recent progress, MLLM-based moderation systems still struggle with several fundamental challenges in real-world moderation tasks.~
\textbf{\emph{Challenge 1}}: Misalignment between model outputs and platform policies. 
Existing models are trained on annotated data, where annotations are based on human interpretations of complex platform guidelines, which are often expressed as intricate textual rules with edge cases and subjective boundaries. Therefore, while the model learns statistical patterns from labels, it lacks direct access to the underlying rules those labels are based on, which often causes the predictions of models to diverge from current moderation expectations. 
\textbf{\emph{Challenge 2}}: Lack of transparency in the decision-making process. Most existing approaches produce black-box predictions without surfacing verifiable evidence or logical explanations~\cite{jhaver2019transparency}, limiting trust and hampering operational collaboration between algorithm teams and human reviewers~\cite{doshi2017towards}. Unable to determine whether the model is reliable and which parts of the system need to be optimized, departments disagreed, resulting in significant communication and operations costs.~
\textbf{\emph{Challenge 3}}: Difficulty in modeling fine-grained label distinctions under complex, ambiguous, and adversarial conditions. Real-world moderation scenarios often involve subtle differences between sibling categories that share higher-level semantics but differ in definition boundaries. Misclassifications frequently occur between such closely related categories, \eg  \emph{Excessive adultification of minors} \emph{vs.} \emph{Inappropriate attire for minors}. 
Worse still, these mistakes between adjacent categories can lead to either over-flagging or under-enforcement, reducing the fairness and reliability of moderation outcomes.~

Together, these challenges create a systemic disconnect between models, human reviewers, and platform governance. Misalignment with moderation rules, lack of transparency, and poor recognition reinforce each other, leading to reduced trust in automated systems, inconsistent enforcement outcomes, and increased human intervention costs~\cite{jo2024harmful}. Addressing these intertwined issues requires a paradigm shift that redefines how decisions are made, explained, and evaluated in content moderation. 

To address these challenges, we propose Hi-Guard, a novel hierarchical content safety framework designed for the decision layer. Hi-Guard incorporates three core innovations.~
\textbf{\emph{Innovation 1}}: Hi-Guard learns from rules, not just data. Instead of learning implicitly from labeled data alone, Hi-Guard explicitly integrates category-level policy definitions into the model prompt. This allows the model to reason over rules at inference time and align its decisions with current platform standards, rather than merely mimicking past annotations. As a result, Hi-Guard can generalize better to unseen cases, reflect the intention of updated policies, and provide traceable justifications for each classification.
\textbf{\emph{Innovation 2}}: 
Hi-Guard introduces an explicit chain-of-thought (CoT)~\cite{wei2022chain} reasoning component, coupled with reinforcement learning from verifiable accuracy and format rewards (RLVR)~\cite{guo2025deepseekr1, team2025kimi}, where the model outputs not only its decision but also a detailed explanation~\cite{kojima2022large}. 
This RLVR-driven policy optimization Group Relative Policy Optimization (GRPO)~\cite{shao2024deepseekmath}, enables Hi-Guard to deliver transparent, auditable, and rule-compliant decisions that can be understood and trusted by human moderators.~
\textbf{\emph{Innovation 3}}: Hierarchical taxonomy and path-based soft reward restraint. We introduce a hierarchical taxonomy in which each category is represented as a structured path spanning domain, topic, subtype, and behavior, mimicking real-world moderation policy granularity. Unlike the binary hard reward in RLVR, a soft-margin reward is introduced to penalize sibling-category confusion more than distant errors, with increasing penalties at deeper levels. This structure-aware design improves the model’s ability to distinguish subtle risks and generalize to ambiguous or unseen cases. 
These key innovations together establish a novel decision paradigm: the decision \& evidence layer. This new layer replaces the original decision layer, enabling models to output not only classification results but also interpretable rationales that align with platform policies, fostering transparency and rule consistency. Extensive offline and online evaluations demonstrate that Hi-Guard, built upon this paradigm, significantly improves classification accuracy, reasoning quality, and system efficiency, effectively breaking the vicious cycle between models, human reviewers, and moderation rules.~
This paper makes the following key contributions:
\begin{itemize}
\item We explicitly incorporate platform-defined rule definitions into model inputs, allowing the model to produce structured, interpretable predictions grounded in real moderation policies.

\item We design a multi-level taxonomy that reflects the hierarchical structure of platform risk definitions, enabling path-based reasoning in a top-down manner and improving generalization to unseen cases.

\item We propose a soft-margin reward function that penalizes sibling-category confusions with increasing depth-aware severity, and optimize the model using GRPO to enhance fine-grained learning and reasoning quality.

\item We present Hi-Guard as a unified policy-aligned decision paradigm that bridges the gap between rule-based governance and data-driven modeling, enabling scalable and trustworthy content moderation.
\end{itemize}
\begin{figure*}[t]
  \includegraphics[width=\linewidth]{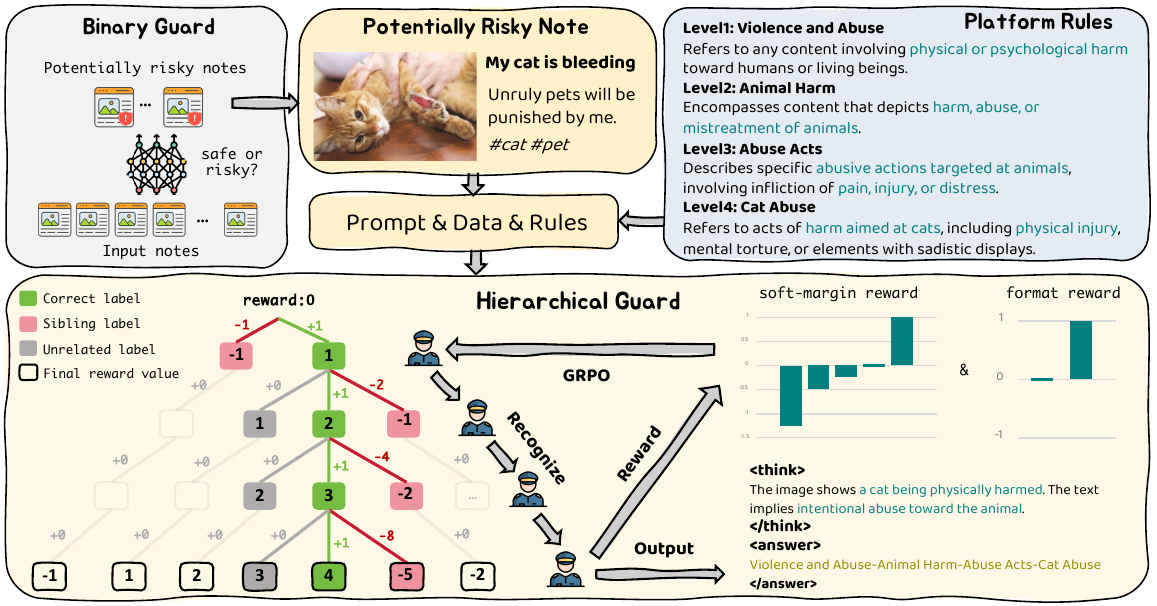}
  \caption{The framework of our proposed Hi-Guard. “Hierarchical” refers to both the two-stage moderation pipeline and the multi-level taxonomy used for path-based classification. 
  The binary guard recognizes the input notes and captures the potentially risky notes, the hierarchical guard then accepts them and platform rules as the input, with format and soft-margin accuracy reward, driving the model to learn semantic structure among labels and search for the correct reasoning path.}
  \label{fig:framework}
\end{figure*}
\section{Related Work}
\textbf{MLLMs for Content Moderation.}~
As content moderation systems evolve into multi-stage pipelines, MLLMs are increasingly deployed in the decision layer to handle ambiguous, high-risk, and policy-sensitive cases.~
Hee \etal~\cite{hee2024recent} point out that while MLLMs improve coverage over both text and image inputs, their decisions often lack explicit alignment with platform-specific policies. To address evaluation gaps, Jin \etal~\cite{jin2024mmsoc} introduce MM-SOC, a benchmark designed to assess MLLMs under real-world social media settings. Their results show that existing models struggle with multimodal reasoning, label generalization, and multilingual content, revealing limitations in robustness and policy alignment. Wu \etal~\cite{wu2025icm} further propose ICM-Assistant. 
By integrating policy documents into the input and requiring structured outputs, their model achieves alignment with moderation guidelines and generates rationale-enhanced predictions. Complementarily, Jo \etal~\cite{jo2024harmful} propose a fine-grained taxonomy of online harms and evaluate MLLMs as alternative annotators for YouTube video moderation. Their findings suggest that MLLMs can approximate human judgments across many harm categories but still fall short in consistent explanation quality.

\textbf{Reinforcement Learning.}~
To align language model behavior with human preference or task-specific constraints, reinforcement learning has emerged as a popular optimization strategy beyond traditional supervised fine-tuning. 
Reinforcement Learning from Human Feedback (RLHF)~\cite{ouyang2022training} has been widely used to train LLMs, where scalar preference scores are used to optimize the model policy~\cite{ziegler2019fine, glaese2022improving,wu2021recursively}.~
Beyond preference comparisons, many approaches transform human feedback, including pairwise judgments, binary correctness, and rule compliance, into reward signals that supervise the training process~\cite{rafailov2023direct, stiennon2020learning, hancock2019learning, zhou2020learning, perez2019finding, sun2023aligning, yu2024rlhf, zhao2023beyond}.~
More recently, RLVR~\cite{guo2025deepseekr1} has been introduced to incorporate structured and interpretable reward signals, such as classification correctness and format consistency~\cite{liu2025visual, yu2025perception}.~
These reward designs often encourage structured outputs via CoT prompting~\cite{wei2022chain, kojima2022large}, enabling models to expose their reasoning process and improving decision transparency.
To further enhance training robustness, GRPO~\cite{shao2024deepseekmath} is introduced, which aggregates policy updates across example groups, improves sample efficiency, and reduces variance under sparse reward conditions.
\section{Methodology}

\subsection{System Overview}
Hi-Guard is a cascaded moderation framework designed to balance classification precision, interpretability, and inference efficiency in real-world content safety systems. As shown in Figure~\ref{fig:framework}, it consists of two core stages: a lightweight binary classifier that filters safe content with high risk recall, and a stronger hierarchical classifier that performs fine-grained multi-risk classification on the remaining potentially risky content. 

This design is motivated by two key observations:~
(1) Determining whether a piece of content is risky or not is itself a highly nontrivial task, often requiring high recall and conservative judgments to prevent harmful content from being overlooked;
(2) Risky content constitutes only a small fraction (typically around 20\% in the decision layer) of all UGC, leading to severe class imbalance. Performing binary risk detection and fine-grained classification in a single-stage model would conflate distinct objectives and degrade overall performance.~
By decoupling these tasks, Hi-Guard simplifies the learning targets of each model and ensures that the heavyweight classifier focuses exclusively on ambiguous or high-stakes decisions.

\subsection{Stage 1: Binary Risk Filtering}
Given a multimodal input note $N = (T, V)$, where $T$ and $V$ denote the textual and visual modalities, respectively, the binary classifier $f_1(\cdot)$, instantiated as Qwen2-VL-2B with parameters $\theta_1$, predicts a coarse-grained binary label:
\begin{equation}
s = f_1(N;\theta_1) \in \{\text{safe, risky}\}
\end{equation}
We fine-tune the binary classifier using supervised fine-tuning (SFT) with a cross-entropy loss:
\begin{equation}
\mathcal{L}_{\text{SFT}} = - \mathbb{E}_{(N, s) \sim \mathcal{D}} \left[ \log P_{\theta_1}(s\mid N) \right]
\end{equation}
where $s \in \{0, 1\}$ is the binary ground-truth label indicating safe (0) or risky (1) content. 

\subsection{Stage 2: Hierarchical Risk Classification}\label{sec:stage2}
The second stage addresses three key limitations in fine-grained risk classification, as discussed in Section~\ref{sec:intro}: (1) misalignment between model outputs and platform policies, (2) lack of interpretability in model decisions, and (3) insufficient discrimination among fine-grained and easily confusable categories.

To mitigate the issue of policy misalignment, we explicitly incorporate platform rule definitions into the model prompt. The classifier receives a structured prompt that includes the candidate categories and their associated policy guidelines. The model is required to output both a reasoning trace and a final decision in a predefined format: \texttt{<think>} CoT reasoning grounded in rules \texttt{</think>} \texttt{<answer>} risk category or "No Risk" \texttt{</answer>}. This design ensures that decisions are traceable, verifiable, and consistent with moderation standards.~

\begin{figure}[t]
\centering
\begin{subfigure}[t]{0.495\linewidth}
  \includegraphics[width=\linewidth]{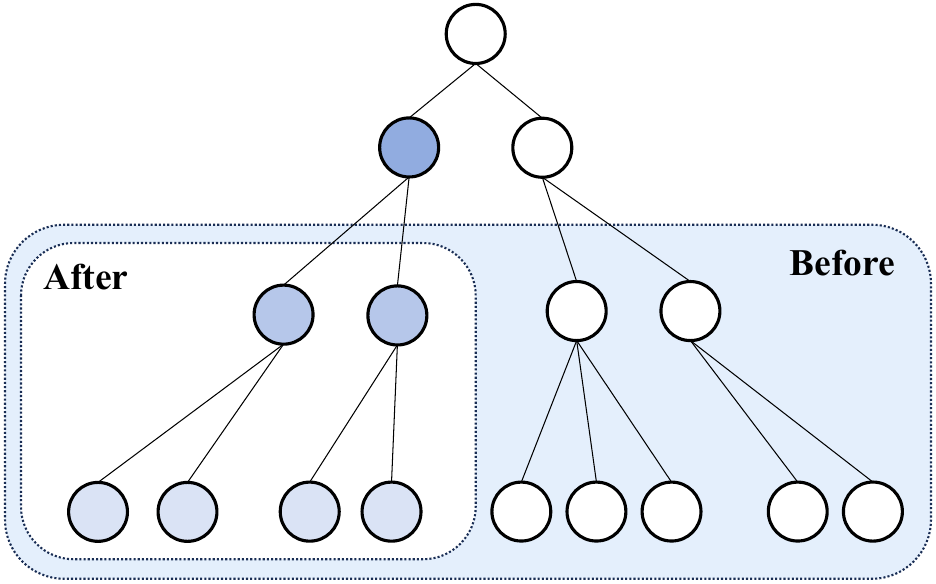}
  \caption{Hierarchical decomposition narrows solution space.}
  \label{fig:amongclass}
\end{subfigure}%
\hfill  
\begin{subfigure}[t]{0.495\linewidth}
  \includegraphics[width=\linewidth]{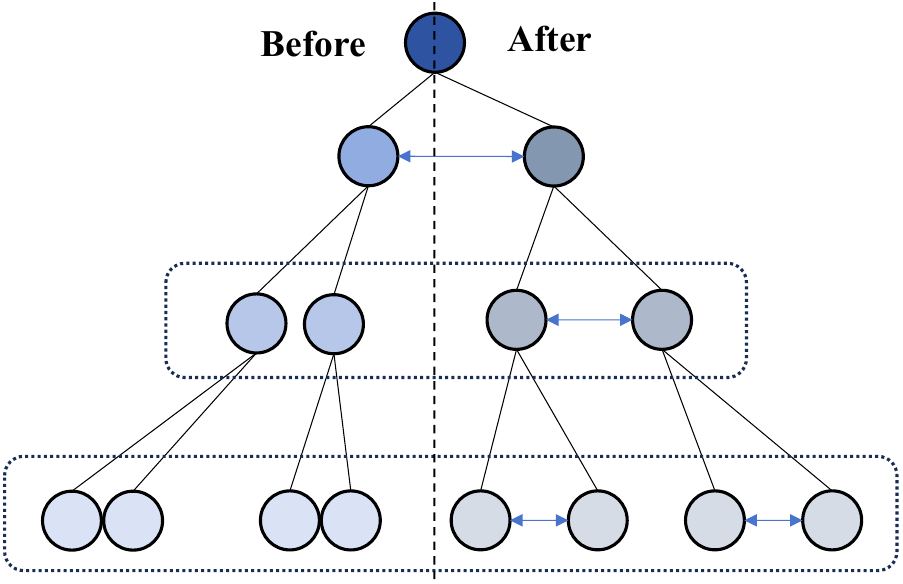}
  \caption{Specifically designed reward reduces sibling confusion.}
  \label{fig:interclass}
\end{subfigure}
\caption{Illustration of Hi-Guard’s structural benefits. (a) The hierarchical taxonomy reduces the prediction space by limiting candidate labels at each level. 
(b) Sibling categories under the same level are more easily confused. Hi-Guard encourages the model to separate such categories more clearly.}
\label{fig:innovations}
\end{figure}

To further improve classification performance and generalization, we introduce two structural innovations, illustrated in Figure~\ref{fig:innovations}. As shown in Figure~\ref{fig:innovations} (a), we construct a hierarchical label taxonomy aligned with real-world moderation policies. Each category is represented as a path consisting of multiple levels of categories. In our implementation, the taxonomy follows a four-level structure: Domain $\rightarrow$ Topic $\rightarrow$ Subtype $\rightarrow$ Behavior. This formulation transforms the flat classification task into a stepwise path prediction problem, where the model progressively selects one label at each level. This coarse-to-fine prediction not only reduces the solution space at each step but also encourages more focused and interpretable predictions.

Figure~\ref{fig:innovations} (b) highlights a common challenge in moderation: confusion among sibling categories at the same level. For example, behavior-level labels under the same subtype often exhibit overlapping definitions or blurred boundaries. To address this, we apply structural supervision that increases the decision margin between such confusable options, encouraging the model to differentiate them more cautiously.

To support these designs, we construct a structured prompt that presents the model with the definitions of candidate categories, their parent class, and siblings at each level, as detailed in Figure~\ref{fig:prompt}. The model output needs to follow: \texttt{<think>} CoT reasoning referencing rules and siblings \texttt{</think>} \texttt{<answer>} Full 4-level risk path or "No Risk" \texttt{</answer>}.~Specifically, only samples classified as risky in the first stage are forwarded to the fine-grained classifier $f_2(\cdot)$, instantiated as Qwen2-VL-7B with parameters $\theta_2$, which outputs the final multi-level risk label:

\begin{equation}
y =
\begin{cases}
\text{safe}, & \text{if } s = \text{safe} \\
f_2(N; \theta_2), & \text{if } s = \text{risky}
\end{cases}
\end{equation}

\begin{figure}[t]
\footnotesize
\begin{tcolorbox}[
    colback=blue!2!white,
    colframe=lightblue!30!black,
    coltitle=white,
    colbacktitle=lightblue!80!white,
    title=Hi-Guard Prompt Template,
    fonttitle=\bfseries,
    boxrule=1pt,
    enhanced
  ]

\textbf{System Prompt} \\[3pt]
\texttt{<|im\_start|>} \texttt{system} \\[2pt]
Given a multi-modal content sample, select the most appropriate risk path from the following four-level taxonomy and provide your reasoning.\\[3pt]

\textcolor{lightblue!80!black}{\textbf{Category Taxonomy \& Rule Definitions:}} \\[3pt]
\textbf{Level-1:} \textit{Minor} — individuals aged between 3 and 18 years old. \\
\textbf{Level-2:} \textit{Inappropriate Behavior Involving Minors} — Refers to behaviors presented by minors in the content that serve as negative examples, induce, imply, or glamorize such actions ... \\
\textbf{Level-3:} \textit{Delinquent Social Atmosphere} — Refers to minors imitating or participating in negative adult social behaviors such as smoking, drinking, partying in nightclubs, fighting, or ... \\
\textbf{Level-4:} \textit{Underage Drinking} — Explicit or implicit depictions of minors consuming alcohol, including ... \\[4pt]

\textcolor{lightblue!80!black}{\textbf{Category Path:}} \\[3pt]
Minor $\rightarrow$ Inappropriate Behavior Involving Minors $\rightarrow$ Delinquent Social Atmosphere $\rightarrow$ Underage Drinking \\[4pt]
...\\

\textcolor{lightblue!80!black}{\textbf{Instructions:}}
\begin{itemize}[leftmargin=1.2em]
    \item Select the most appropriate risk path. If no risk, answer No Risk.
    \item Explain why you chose this category using definitions of the selected category and its siblings. Be specific.
    \item Output format: \\
    \texttt{<think>} Reasoning ($\geq$ 20 words) \texttt{</think>} \texttt{<answer>} Full risk path or \texttt{No Risk} \texttt{</answer>}
\end{itemize}
\texttt{<|im\_end|>} \\[3pt]

\textbf{User Prompt} \\[3pt]
\texttt{<|im\_start|>} \texttt{user} \\
Please give categorical judgment and reasoning for the following: \\
\textbf{Content}: \{Image + Text\} \\
\texttt{<|im\_end|>}

\end{tcolorbox}
\caption{Illustration of the structured prompt used in Hi-Guard.}
\label{fig:prompt}
\end{figure}

\subsection{Reward Design and GRPO Optimization}
We adopt GRPO to train the multi-level classifier in the second stage.~
This approach allows us to provide feedback beyond simple correctness and encourages structured, policy-aligned outputs. We design a composite reward that guides the model to produce both accurate and interpretable predictions. 

\textbf{Format Reward.}~
We encourage structured output: \texttt{<think>} ... \texttt{</think>} \texttt{<answer>} ... \texttt{</answer>}, which separates the reasoning process from the final decision and supports better human auditability. The format reward is defined as:
\begin{equation}
R_{\text{format}} =
\begin{cases}
1, & \text{if output matches the format} \\
0, & \text{otherwise}
\end{cases}
\end{equation}

\textbf{Path-aware Soft-margin Accuracy Reward.}~
As discussed in Section~\ref{sec:stage2}, sibling categories under the same parent are often more confusable due to subtle or overlapping definitions. To encourage the model to better distinguish these closely related classes, we replace conventional binary feedback with a graded reward scheme that reflects the structural relationship between predicted and ground-truth categories.~
This reward design distinguishes three cases at each level: correct predictions receive a reward; sibling-category errors are penalized with negative values; unrelated mistakes receive zero reward. The penalties grow exponentially with depth using $-2^{l-1}$, assigning stronger feedback to errors at finer-grained levels, reflecting their greater importance and difficulty in real-world moderation.~
Let the predicted category path be $\hat{P} = (\hat{y}^{(1)}, \hat{y}^{(2)}, \dots, \hat{y}^{(L)})$ and the ground-truth path be $P = (y^{(1)}, y^{(2)}, \dots, y^{(L)})$. The per-level reward is:
\begin{equation}
R_{\text{acc}}^{(l)} =
\begin{cases}
1, & \text{if } \hat{y}^{(l)} = y^{(l)} \\
-2^{l-1}, & \text{if } \hat{y}^{(l)} \in \text{sibling}(y^{(l)}) \\
0, & \text{otherwise}
\end{cases}
\end{equation}
In addition, we assign the most severe negative reward to penalize top-level misclassifications between safe and risky content.~
All reward configurations are detailed in Appendix~\ref{supp:reward}.~
The final reward aggregates accuracy and format components:
\begin{align}
R_{\text{acc}} &= \frac{1}{L} \sum_{l=1}^{L} R_{\text{acc}}^{(l)} \\
R_{\text{final}} &= R_{\text{acc}} + R_{\text{format}}
\end{align}

\textbf{GRPO Training Objective.}~
\begin{table}[t]
    \caption{Detailed statistics of our dataset.}
    \centering
    \begin{tabular}{lcl}
    \toprule
    \multicolumn{3}{c}{\textbf{Stage 1: Binary Classification}} \\
    \midrule
    Split & \#Classes & Notes (Risky / Safe) \\
    \midrule
    Training & 2 & 28,000 (20,000 / 8,000) \\
    Evaluation & 2 & 10,000 (2,000 / 8,000) \\
    \midrule
    \midrule
    \multicolumn{3}{c}{\textbf{Stage 2: Multi-class Classification}} \\
    \midrule
    Split & \#Classes & Notes (Risky / Safe) \\
    \midrule
    Training & 20 & 24,000 (20,000 / 4,000) \\
    Base Eval. & 20 & 10,000 (2,000 / 8,000) \\
    Generalization Eval. & 14 & 7,000 (1,400 / 5,600) \\
    \bottomrule
    \end{tabular}
    \label{tab:dataset}
\end{table}
GRPO generates a group of $G$ responses for each input note $N$, scores them using a composite reward $R_{\text{final}}$, and applies group-wise normalization for stable learning.~
Given $G$ response sequences $\{\hat{P_i}\}_{i=1}^G$ generated by the model, each with reward $R_i$, GRPO computes the normalized advantage for the $i$-th sample as:
\begin{equation}
A_i = \frac{R_i - \text{mean}(R_1, R_2, \dots, R_G)}{\text{std}(R_1, R_2, \dots, R_G)}.
\label{eq:grpo-advantage}
\end{equation}

The training objective is to maximize the expected log-probability of the full category path $\hat{P}$, weighted by the normalized advantage:
\begin{equation}
\mathcal{L}_{\text{GRPO}} = \mathbb{E}_{N \sim \mathcal{D}} \left[ A_i \cdot \log \pi_{\theta_2}(\hat{P}_i\mid N) \right].
\label{eq:grpo-loss}
\end{equation}
Here, $\pi_{\theta_2}$ denotes the classification policy and $A_i$ is the relative quality of the $i$-th sample within the group. This formulation encourages the model to prefer higher-quality outputs while avoiding reliance on an explicit critic.

\section{Experiments}
We conduct comprehensive experiments to evaluate the effectiveness of Hi-Guard in terms of classification accuracy, generalization, interpretability, and deployment efficiency. Our study addresses the following research questions:
\begin{itemize}[leftmargin=1.2em]
\item \textbf{RQ1:} Does Hi-Guard outperform baseline MLLMs in overall moderation performance under real-world conditions?
\item \textbf{RQ2:} How do policy introduction, hierarchical labels, and soft-margin reward contribute to performance gains?
\item \textbf{RQ3:} Can Hi-Guard demonstrate true policy understanding and maintain robustness on real-world moderation data, thereby reflecting its potential as a reliable safety moderation model?
\end{itemize}
\subsection{Dataset and Experiment Setup}

\textbf{Data Preprocessing.} To ensure consistent multimodal input formatting, we filter out incomplete samples (\eg notes missing image or text), and truncate long texts to 512 tokens. Notes are de-duplicated and anonymized to preserve privacy. 

\begin{table*}[t]
    \caption{Zero-shot performance comparison across models on the generalization set. We report overall accuracy and average precision/recall on risky categories.}
    \begin{center}
    \begin{tabular}{l|c|ccc}
    \toprule
    \textbf{Method} & \textbf{\#Params} & \textbf{Overall Accuracy} & \textbf{Precision} & \textbf{Recall} \\
    \midrule
    Qwen2-VL & 7B & 71.09\% & 36.83\% & 27.93\% \\
    Qwen2-VL + SFT & 7B & 71.98\% & 38.70\% & 49.14\% \\
    \rowcolor[HTML]{DAEFF9}
    Hi-Guard \textbf{[Ours]} & 7B & \textbf{84.11\%} & \textbf{52.72\%} & \textbf{59.42\%} \\
    \textbf{\makebox[0.13\linewidth][c]{$\Delta$ (vs. SFT)}} & - & \hgreen{+12.13\%} & \hgreen{+14.02\%} & \hgreen{+10.28\%} \\
    \bottomrule
    \end{tabular}
    \end{center}
    \label{tab:mllm-result}
    \vspace{0.09cm}
\end{table*}
\begin{table*}[t]
    \caption{Ablation study on key components of Hi-Guard. We report overall accuracy and average precision/recall on risky categories from both the base and generalization evaluation sets. We report the total GPU inference time on 8 H800 GPUs across both base and generalization sets.}
    \begin{center}
    \begin{tabular}{l|ccc|ccc|c}
    \toprule
    \multirow{2}{*}{\textbf{Model Variant}} & \multicolumn{3}{c|}{\textbf{Base Evaluation}} & \multicolumn{3}{c|}{\textbf{Generalization Evaluation}} & \multirow{2}{*}{\textbf{GPU Time (h)}} \\
    \cmidrule{2-4} \cmidrule{5-7}
     & \textbf{Overall Acc.} & \textbf{Precision} & \textbf{Recall} & \textbf{Overall Acc.} & \textbf{Precision} & \textbf{Recall} \\
    \midrule
    \multicolumn{8}{c}{\textit{Single-stage (7B)}} \\
    \midrule
    RLVR \textbf{[Base]} & 72.82\% & 43.38\% & 72.45\% & 62.68\% & 34.57\% & 52.78\% & 1.10 \\
    \quad + Platform Rules & 75.84\% & 47.84\% & 77.95\% & 73.81\% & 38.40\% & 53.71\% & 1.61 \\
    \quad\quad + Hierarchical Labels & 79.23\% & 53.67\% & 76.60\% & 80.59\% & 49.70\% & 51.86\% & 2.76 \\
    \quad\quad\quad + Soft-margin Reward & 81.22\% & 52.47\% & 79.35\% & 80.66\% & 48.68\% & 57.79\% & 2.06 \\
    \midrule
    \multicolumn{8}{c}{\textit{Cascaded (2B + 7B)}} \\
    \midrule
    Stage 1 + RLVR & 83.28\% & 52.63\% & 73.44\% & 79.16\% & 42.49\% & 54.23\% & 0.53 \\
    \quad + Platform Rules & 84.33\% & 55.90\% & 78.68\% & 81.56\% & 44.06\% & 54.65\% & 0.74 \\
    \quad\quad + Hierarchical Labels & 85.57\% & 60.50\% & 77.75\% & 83.80\% & 53.11\% & 53.12\% & 1.27 \\
    \rowcolor[HTML]{DAEFF9}
    \quad\quad\quad + Soft-margin Reward \textbf{[Hi-Guard]} & \textbf{86.52\%} & \textbf{59.30\%} & \textbf{80.47\%} & \textbf{84.11\%} & \textbf{52.72\%} & \textbf{59.42\%} & \textbf{0.85} \\
    \textbf{\makebox[0.3\linewidth][c]{$\Delta$ (vs. Base)}} & \hgreen{+13.70\%} & \hgreen{+15.92\%} & \hgreen{+8.02\%} & \hgreen{+21.43\%} & \hgreen{+18.15\%} & \hgreen{+6.64\%} &  \hgreen{-22.73\%} \\
    \bottomrule
    \end{tabular}
    \end{center}
    \label{tab:ablation-result}
\end{table*}
\textbf{Dataset Construction.}~We construct a two-stage dataset of multimodal notes collected from a real-world content moderation platform. Stage1 includes binary safe/risky labels, while stage2 provides structured category paths. To evaluate both in-domain accuracy and generalization capability, we divide the evaluation data in stage2 into a base and a generalization set, which contain fully observed during training and previously unseen categories, respectively.~

\textbf{Training and Evaluation.} All models are trained for one epoch with SFT or GRPO. In the evaluation phase, we report: (1) overall classification accuracy, precision, and recall on risky classes; (2) representative model outputs demonstrating reasoning and prediction structure; and (3) human preference rate from a reasoning quality ranking study. The training hyperparameters and the structure of evaluation sets are described in Appendix~\ref{supp:train_and_eval}.

\subsection{Main Results}


\textbf{Compare with Baselines.}~
We select Qwen2-VL as the base model for comparison due to its strong open-source performance in multimodal understanding and its suitability for instruction tuning. As a publicly available MLLM, it provides a transparent and reproducible baseline. The SFT variant represents a straightforward supervised adaptation to our moderation tasks and serves as a competitive baseline.~
As illustrated in Table~\ref{tab:mllm-result}, our Hi-Guard consistently outperforms both the base Qwen2-VL model and its supervised fine-tuned variant across all metrics. Compared to SFT, Hi-Guard yields more accurate predictions on unseen risky subcategories where the understanding of data and rules is critical. This performance gain demonstrates the effectiveness and robustness of our Hi-Guard under real-world moderation conditions.

\textbf{Ablation Study.}~
To better understand the contributions of individual components, we perform an ablation study based on two architectural configurations: the single-stage pipeline and the cascaded pipeline. The base model is Qwen2-VL trained with RLVR, and we progressively add the following components: (1) Platform Rules: injecting category-specific definitions into prompts; (2) Hierarchical Labels: replacing flat labels with a four-level category path; and (3) Soft-Margin Reward: a structure-aware accuracy signal that penalizes sibling-category confusion more than distant errors.

Table~\ref{tab:ablation-result} summarizes the results on both the base and generalization evaluation sets. We observe substantial improvements across most metrics as each component is added. Notably, moving from single-stage to cascaded architecture not only improves performance but also significantly reduces GPU cost. Our final model, Hi-Guard, achieves the best overall accuracy, precision, and recall, while reducing GPU time by 22.7\% compared to the single-stage base model.~

These results demonstrate that our design choices, including rule grounding, hierarchical labels, and reward shaping, not only enhance classification robustness and interpretability but also bring practical efficiency advantages for real-world deployment.

\begin{table*}[th]
\caption{A case study demonstrating fine-grained classification capability of Hi-Guard and its cautious perception of risk. \textcolor{olive}{Olive} text denotes the correct output and \textcolor{teal}{Teal} text denotes the thought of the model Hitting Rules.}
\footnotesize
\setlength\tabcolsep{4pt}
\centering
\begin{tabular} {p{1.8cm}|p{15.4cm}}
    \toprule
    \multirow{7}{*}{\textbf{Rule}} & \textit{\textbf{Minor}: Individuals aged between 3 and 18 years old.} \\
    & \textit{\textbf{Inappropriate Behavior Involving Minors}: Refers to behaviors presented by minors in the content that serve as negative examples, induce, imply, or glamorize such actions, potentially misleading the audience, encouraging imitation, or distorting values.} \\
    & \textit{\textbf{Delinquent Social Atmosphere}: \textcolor{teal}{Refers to minors imitating or participating in negative adult social behaviors such as smoking}, \textcolor{teal}{drinking}, partying in nightclubs, fighting, or getting tattoos—showing signs of premature maturity or risky tendencies.} \\
    & \textit{\textbf{Underage Drinking}: Explicit or implicit depictions of minors consuming alcohol, \textcolor{teal}{such as holding beer bottles, drinking alcoholic beverages, or being in alcohol-related settings}.} \\
    \midrule
    \textbf{Note} &
    \begin{minipage}[c]{\linewidth}
        \begin{minipage}[c]{0.15\linewidth}
            \includegraphics[width=\linewidth]{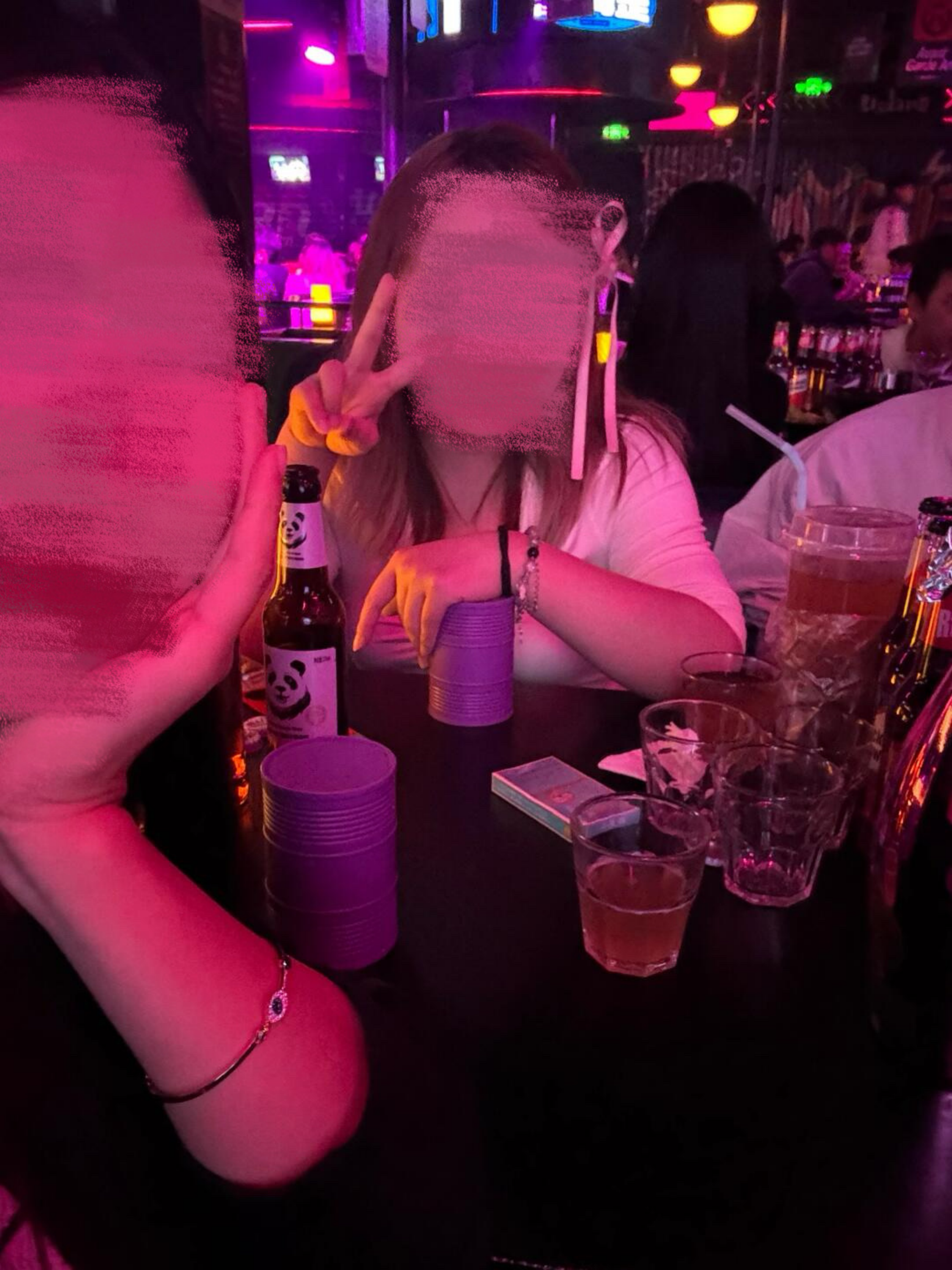}
        \end{minipage}
        \hspace{0.8em}
        \begin{minipage}[c]{0.7\linewidth}
            \textbf{[Title]}: \textit{Let’s Go! Bar Party}

            \textbf{[Content]}: \textit{First time sneaking into a bar and trying vodka with friends \\ We told them we were 20 but nobody checked \\ Honestly, the mojito was super sweet-didn't even taste like alcohol! \\ \#FirstBarExperience \#WorthIt}
            
            \textbf{[OCR Content]}: \textit{empty}
        \end{minipage}
    \end{minipage}
    \\
    \midrule
    \textbf{Ground Truth} & Minor–Inappropriate Behavior Involving Minors–Delinquent Social Atmosphere–Underage Drinking \\
    \midrule
    \multirow{1}{*}{SFT} & \textcolor{olive}{Underage Drinking} \\
    \midrule
    \multirow{2}{*}{RLVR} & \textit{<think>} In the image, \textcolor{teal}{two girls are drinking beverages in a bar, and one of them is holding a bottle of beer}. \textit{</think>} \\
    & \textit{<answer>} \textcolor{olive}{Underage drinking} \textit{</answer>} \\
    \midrule
    \multirow{5}{*}{Hi-Guard} & \textit{<think>} \textcolor{teal}{The image shows two girls sitting in a bar, with one of them drinking beer and making a victory gesture}, suggesting they might be enjoying nightlife or celebrating an event. However, \textcolor{teal}{since drinking alcohol may be considered inappropriate behavior}, we need to further analyze whether there are more specific inappropriate actions or other potential risk factors involved. Based on the provided information, the image does not involve any other illegal or harmful content. \textit{</think>} \\
    & \textit{<answer>} \textcolor{olive}{Minor–Inappropriate Behavior Involving Minors–Delinquent Social Atmosphere–Underage Drinking} \textit{</answer>} \\
    \bottomrule
\end{tabular}
\label{tab:case1}
\end{table*}
\begin{table*}[h!]
\caption{A case study illustrating superior rule comprehension and analytical reasoning of Hi-Guard in ambiguous scenarios. \textcolor{olive}{Olive} text denotes the correct output and \textcolor{teal}{Teal} text denotes the thought of the model Hitting Rules.}
\footnotesize
\setlength\tabcolsep{4pt}
\centering
\begin{tabular} {p{1.8cm}|p{15.4cm}}
    \toprule
    \multirow{5}{*}{\textbf{Rule}} & \textit{\textbf{Minor}: Individuals aged between 3 and 18 years old.} \\
    & \textit{\textbf{Vulgar Content Involving Minors}: The content features minors engaging in vulgar, provocative, flirtatious, \textcolor{teal}{revealing}, or indecent expressions, presenting a clear misguidance in aesthetic orientation or causing uncomfortable associations through visual or verbal presentation.} \\
    & \textit{\textbf{Minor in Inappropriate Attire}: \textcolor{teal}{Content featuring minors wearing revealing clothing}, exposing sensitive body parts, or dressed in a sexually suggestive manner.} \\
    & \textit{\textbf{Underwear Exposure}: \textcolor{teal}{Minors shown wearing only underwear or experiencing accidental exposure of underwear}.} \\
    \midrule
    \textbf{Note} &
    \begin{minipage}[c]{\linewidth}
        \begin{minipage}[c]{0.25\linewidth}
            \includegraphics[width=\linewidth]{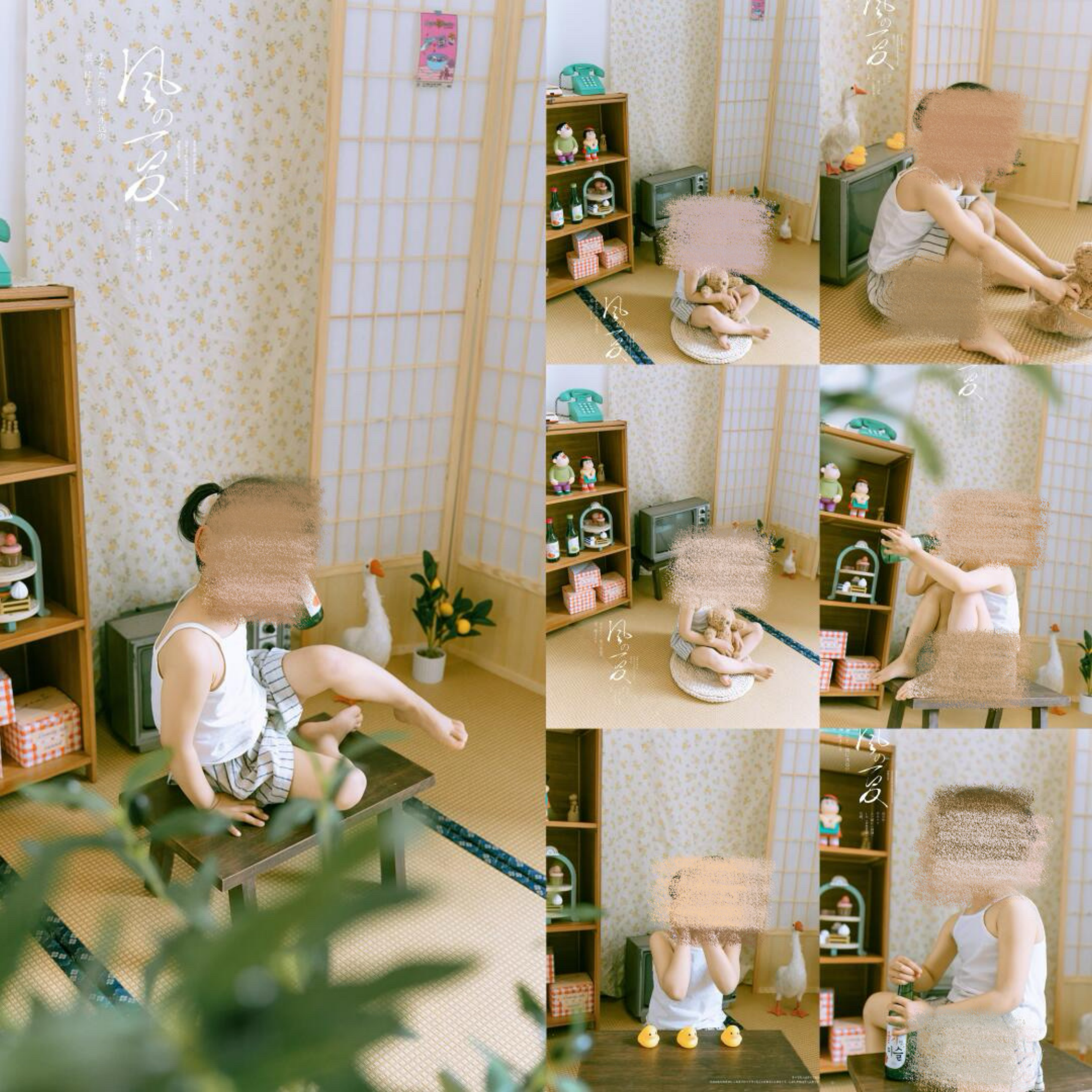}
        \end{minipage}
        \hspace{0.8em}
        \begin{minipage}[c]{0.7\linewidth}
            \textbf{[Title]}: \textit{Dear little one}

            \textbf{[Content]}: \textit{Behind every frame is a gentle narration of your journey growing up. \\So, I’m looking forward to the day, many years from now, when we flip through these moments of happiness together. \\ \#MyGirl \#MomentsToRemember \#ChildPhotography}

            \textbf{[OCR Content]}: \textit{Summer Breeze}
        \end{minipage}
    \end{minipage}
    \\
    \midrule
    \textbf{Ground Truth} & Minor–Vulgar Content Involving Minors–Minor in Inappropriate Attire–Underwear Exposure \\
    \midrule
    \multirow{1}{*}{SFT} & Underage Drinking\\
    \midrule
    \multirow{4}{*}{RLVR} & \textit{<think>} The image shows several children in an indoor photo shoot. One child is sitting on a stool, another is sitting while holding a toy doll, and a third is also sitting while hugging a toy doll. These photos do not display any explicit sexual or violent scenes, nor do they involve political, religious, or other sensitive topics. \textit{</think>} \\
    & \textit{<answer>} \textcolor{olive}{Underwear exposure} \textit{</answer>} \\
    \midrule
    \multirow{7}{*}{Hi-Guard} & \textit{<think>} The image features several children, including a young girl sitting on the floor next to a bottle of alcohol, which may raise concerns about underage drinking. Additionally, \textcolor{teal}{some of the children are wearing tank tops and shorts, which could lead some viewers to perceive their attire as overly revealing. However, under the classification “Minor–Vulgar Content Involving Minors–Minor in Inappropriate Attire–Underwear Exposure,” the key consideration is whether the clothing is age-appropriate, rather than labeling tank tops and shorts as problematic by default}. Therefore, in this case, it is unclear whether any actual risk factors are present. That said, given the possibility of ambiguous cases or special contexts, we should approach the situation with caution and take appropriate measures. \textit{</think>} \\
    & \textit{<answer>} \textcolor{olive}{Minor–Vulgar Content Involving Minors–Minor in Inappropriate Attire–Underwear Exposure} \textit{</answer>} \\
    \bottomrule
\end{tabular}
\label{tab:case2}
\end{table*}

\textbf{Qualitative Comparison.}~
To further illustrate the advantages of Hi-Guard in precise classification and policy-aligned reasoning, we present two qualitative case studies in Table~\ref{tab:case1} and Table~\ref{tab:case2}.~
In the first case, the input depicts two young girls in a bar, with a bottle of beer placed on the table. The SFT baseline provides only a flat label prediction without explanation. While the RLVR model produces a structured output, its reasoning remains superficial, focusing on surface-level image content without linking to platform policies. In contrast, Hi-Guard accurately identifies the full risk path and supports its decision with policy-aware reasoning that explicitly references the behavioral context and risk semantics.~

In the second case, the content depicts a minor in a studio photography setting, wearing light summer clothing. Interestingly, this scene could be misinterpreted in multiple ways: the SFT baseline incorrectly classifies it as "Underage Drinking", likely due to over-reliance on object-level features such as the bottle and gesture, without considering the unlikely nature of actual alcohol consumption in a posed portrait session.~
While the RLVR model outputs the correct label, its reasoning remains superficial and disjointed. It focuses on generic descriptions (\eg children holding dolls) and fails to connect its observations to specific rule definitions. Moreover, its \texttt{<think>} section lacks a clear logical bridge to the final \texttt{<answer>}, resulting in a mismatch between evidence and conclusion.~
By contrast, Hi-Guard not only outputs the correct category path but also demonstrates cautious, rule-aware reasoning. It explicitly evaluates whether the attire and context align with the platform standards, acknowledges the ambiguity, and justifies the need for prudence. This highlights Hi-Guard’s superior capacity to handle borderline cases with interpretability and policy fidelity.

\textbf{Human Evaluation of CoT Quality.}~
To evaluate the quality of model-generated reasoning, we construct a CoT evaluation set containing 170 correctly predicted samples, evenly drawn from both the base and generalization sets across different categories. Each sample includes the original multimodal note, the ground-truth label, and CoT explanations generated by three models: RLVR, Hi-Guard without soft-margin reward (Hi-Guard w/o SMR), and Hi-Guard. These outputs are randomly grouped and evaluated by five professional moderators.

\begin{figure}[t]
  \includegraphics[width=\linewidth]{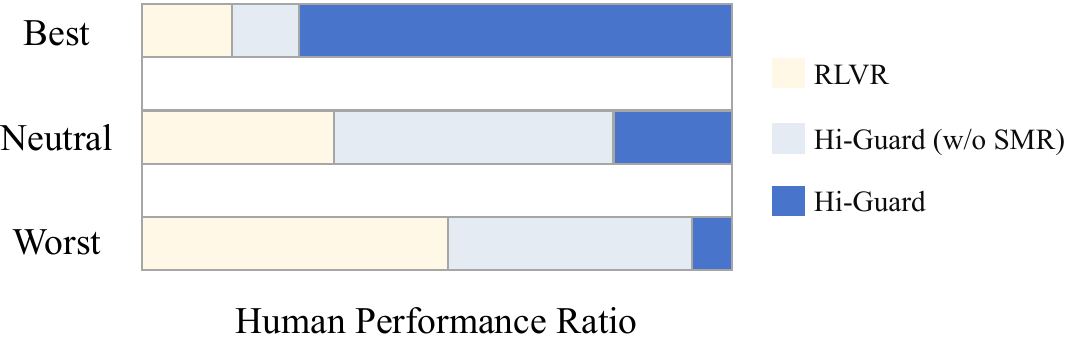}
 \caption{Human preference distribution over CoT explanations from three models (RLVR, Hi-Guard w/o SMR, Hi-Guard) across 170 evaluation samples. Bars represent the average proportion of times each model is rated as best, neutral, or worst by five professional moderators.}
  \label{fig:user_study}
\end{figure}

For each case, moderators select the best, neutral, and worst reasoning among the three. As shown in Figure~\ref{fig:user_study}, Hi-Guard is rated as the best in 73.3\% of cases on average, significantly outperforming the other two models in human preference. In contrast, RLVR and Hi-Guard w/o SMR receive the best ratings in only 15.4\% and 11.3\% of cases, respectively, with RLVR ranked as worst in 52.7\% of cases. These results highlight that Hi-Guard not only improves classification performance but also produces more interpretable and human-aligned reasoning.

\subsection{Online Deployment}
\begin{figure}[h!]
  \includegraphics[width=\linewidth]{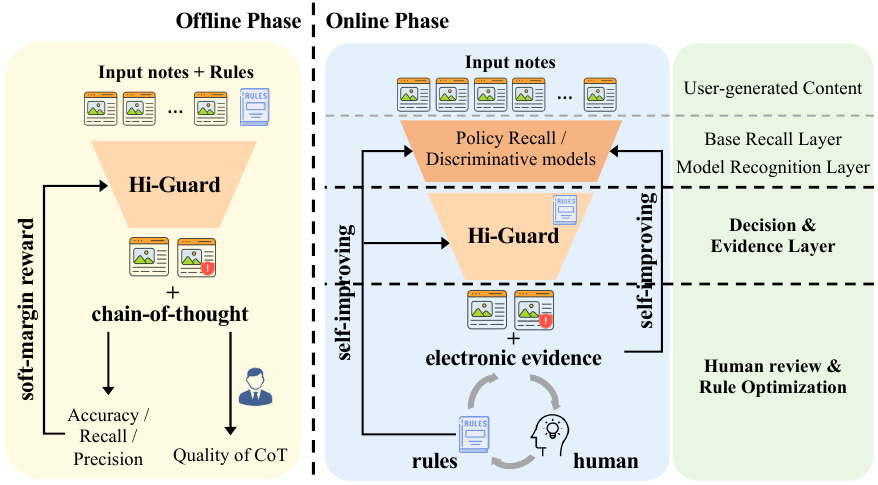}
  \caption{The offline training and online deployment architecture. Compared with the original system architecture, Hi-Guard upgrades the original decision layer into the decision \& evidence layer. Model outputs and human feedback form a closed loop for dynamic rule refinement, enabling policy-aligned, interpretable, and self-improving moderation.}
  \label{fig:online}
\end{figure}

As the final step of automated moderation before human involvement, Hi-Guard plays a crucial role in generating accurate classification results along with structured electronic evidence to support human analysis and reexamination. Figure~\ref{fig:online} illustrates our real-world online deployment architecture, which consists of an offline knowledge alignment phase and a multi-stage online inference pipeline.~
In the offline phase, platform policies and expert-defined moderation rules are integrated into Hi-Guard through explicit input. 
In the online phase, human feedback and model outputs jointly support dynamic rule adjustment and reward calibration. 
This feedback loop enables Hi-Guard to evolve towards near-autonomous moderation with minimal human intervention, reducing the involvement of human moderators and ultimately lowering the operational cost of manual moderation. 

Specifically, we sample 10\% of the online content over a 10-day period, all of which is processed through the complete online moderation pipeline. Hi-Guard achieves an overall accuracy of 85.06\%, an average precision of 51.09\%, and a recall of 79.14\% on risky content.~
In terms of human review efficiency, Hi-Guard reduces the overall human moderation rate by 56.38\%, bringing the final manual review proportion down to just 0.24\%.~
Overall, Hi-Guard strikes a practical balance between high recall of risky content and low manual dependency. These results highlight Hi-Guard’s potential as a reliable, explainable, and cost-effective solution for real-world content safety systems.
\section{Conclusion}
We present Hi-Guard, a cascaded moderation framework for hierarchical risk classification in multimodal content. By integrating explicit policy definitions into prompt design and optimizing with GRPO, Hi-Guard aligns model behavior with evolving moderation standards. Its two-stage architecture combines efficient binary filtering with path-based fine-grained classification, enabling accurate and interpretable decisions across ambiguous and emerging risks. Through structured reasoning and soft-margin reward, Hi-Guard effectively resolves sibling-category confusion while maintaining high recall and auditability. Comprehensive evaluations confirm its improvements over strong MLLM baselines in terms of accuracy, generalization, and system efficiency. Our results highlight the effectiveness of policy-aligned, reward-driven modeling for building reliable and explainable moderation systems in high-stakes, real-world environments.

\begin{acks}
This work was supported by the National Key Research and Development Program of China under Grant 2024YFF0509700,  National Natural Science Foundation of China (62471290, 62331014), and the Fundamental Research Funds for the Central Universities.
\end{acks}

\clearpage
\bibliographystyle{ACM-Reference-Format}
\balance
\bibliography{main}

\newpage
\appendix
\section*{Appendix}
\section{Training \& Evaluation Details}\label{supp:train_and_eval}
We detail the training and evaluation configurations used in our experiments in this section. 

\subsection{Training Configurations}
We summarize the hyperparameter settings for both SFT and GRPO training in Tables~\ref{tab:sft-hyper} and~\ref{tab:grpo-hyper}. 
\begin{table}[!h]
    \centering
    \caption{Detailed training hyperparameters for SFT.}
\begin{tabular}{lc}
\toprule
\textbf{Configuration} & \textbf{Hyperparameter} \\
\midrule
Base model & Qwen2-VL-7B-Instruct \\
Optimizer & AdamW \\
Learning rate & $1 \times 10^{-4}$ \\
Scheduler & Cosine decay \\
Precision & bf16 \\
Gradient checkpointing & false \\
Attention implementation & FlashAttention-2 \\
Deepspeed config & DeepSpeed ZeRO-3 \\
Batch size per device & 8 \\
Gradient accumulation & 1 \\
Global batch size & 8 \\
Epochs & 1 \\
Max pixels  & 802816 \\
LoRA rank & 64 \\
LoRA alpha & 16 \\
LoRA dropout & 0.05 \\
LoRA target & all \\
Warmup ratio & 0.1 \\
Enable liger kernel & true \\
\bottomrule
\end{tabular}
\label{tab:sft-hyper}
\end{table}
\begin{table}[!h]
    \centering
    \caption{Detailed training hyperparameters for GRPO.}
\begin{tabular}{lc}
\toprule
\textbf{Configuration} & \textbf{Hyperparameter} \\
\midrule
Base model & Qwen2-VL-7B-Instruct \\
Optimizer & AdamW \\
Learning rate & $5 \times 10^{-7}$ \\
Scheduler & Cosine decay \\
Precision & bf16 \\
Gradient checkpointing & true \\
Attention implementation & FlashAttention-2 \\
Deepspeed config & DeepSpeed ZeRO-3 \\
Batch size per device & 1 \\
Gradient accumulation & 2 \\
Num. generations & 4 \\
Global batch size & 8 \\
Epochs & 1 \\
Max pixels & 802816 \\
\bottomrule
\end{tabular}
\label{tab:grpo-hyper}
\end{table}
\subsection{Prompt Design and Templates}
To guide the model toward structured and interpretable outputs, we design instruction-based prompts for all phases. The prompt of our Hi-Guard (shown in Figure~\ref{fig:prompt}) incorporates the hierarchical taxonomy and policy definitions.~

Figure~\ref{fig:prompt1} illustrates the baseline RLVR system prompt template, where the model selects a risk category and provides an explanation. Figure~\ref{fig:prompt2} shows the extended version, which includes the policy definitions of each fine-grained category. 
Both prompts enforce a format that separates the model's reasoning (\texttt{<think>...</think>}) from its decision (\texttt{<answer>...</answer>}), enabling interpretability and structured evaluation.
\begin{figure}[h]
\footnotesize
\begin{tcolorbox}[
    colback=blue!2!white,
    colframe=lightblue!30!black,
    coltitle=white,
    colbacktitle=lightblue!80!white,
    title=RLVR System Prompt Template,
    fonttitle=\bfseries,
    boxrule=1pt,
    enhanced
  ]

\textbf{System Prompt} \\[3pt]
\texttt{<|im\_start|>} \texttt{system} \\[2pt]
Given a multi-modal content sample, select the most appropriate risk category from the following categories and provide your reasoning.\\[3pt]

\textcolor{lightblue!80!black}{\textbf{Category:}} \\[3pt]
\textit{Underage Drinking} \\
\textit{Underwear Exposure} \\
...\\

\textcolor{lightblue!80!black}{\textbf{Instructions:}}
\vspace{-1pt}
\begin{itemize}[leftmargin=1.2em]
    \item Select the most appropriate risk category. If no risk, answer No Risk.
    \item Explain why you chose this category. Be specific.
    \item Output format: \\
    \texttt{<think>} Reasoning ($\geq$ 20 words) \texttt{</think>} \texttt{<answer>} Risk Category or \texttt{No Risk} \texttt{</answer>}
\end{itemize}
\texttt{<|im\_end|>}
\end{tcolorbox}
\caption{Illustration of the system prompt of the RLVR base model.}
\label{fig:prompt1}
\end{figure}
\begin{figure}[!h]
\footnotesize
\begin{tcolorbox}[
    colback=blue!2!white,
    colframe=lightblue!30!black,
    coltitle=white,
    colbacktitle=lightblue!80!white,
    title=RLVR (w/ Platform Rules) System Prompt Template,
    fonttitle=\bfseries,
    boxrule=1pt,
    enhanced
  ]

\textbf{System Prompt} \\[3pt]
\texttt{<|im\_start|>} \texttt{system} \\[2pt]
Given a multi-modal content sample, select the most appropriate risk category from the following categories and provide your reasoning.\\[3pt]

\textcolor{lightblue!80!black}{\textbf{Category \& Rule Definitions:}} \\[3pt]
\textit{Underage Drinking} — Explicit or implicit depictions of minors consuming alcohol, including ... \\[4pt]
...\\

\textcolor{lightblue!80!black}{\textbf{Instructions:}}
\vspace{-1pt}
\begin{itemize}[leftmargin=1.2em]
    \item Select the most appropriate risk category. If no risk, answer No Risk.
    \item Explain why you chose this category using definitions of the selected category. Be specific.
    \item Output format: \\
    \texttt{<think>} Reasoning ($\geq$ 20 words) \texttt{</think>} \texttt{<answer>} Risk Category or \texttt{No Risk} \texttt{</answer>}
\end{itemize}
\texttt{<|im\_end|>}
\end{tcolorbox}
\caption{Illustration of the system prompt of the RLVR model with platform rule definitions.}
\label{fig:prompt2}
\end{figure}
\begin{figure*}[t]
  \includegraphics[width=0.9\linewidth]{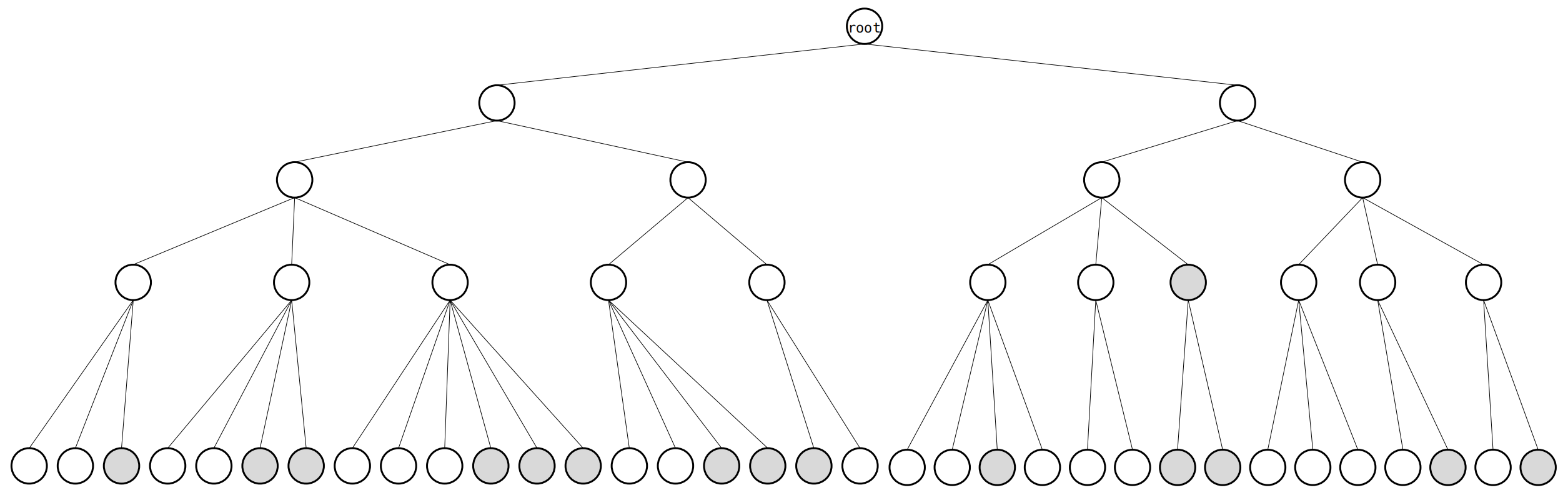}
  \caption{A demonstration of the structure of the evaluation set. The white circles represent known categories present in the training set, and the gray circles represent unseen categories.}
  \label{fig:evaluation-tree}
\end{figure*}
\begin{table*}[h]
\centering
\caption{Possible reward values under different prediction cases in a four-level hierarchical path.
Each $R_{\text{acc}}$ is computed as the average of level-wise rewards, and $R_{\text{final}} = R_{\text{acc}} + R_{\text{format}}$.
}
\label{tab:reward-space}
\begin{tabular}{cccccccccl}
\toprule
\textbf{L1} & \textbf{L2} & \textbf{L3} & \textbf{L4} & \textbf{Format} & 
\textbf{$\boldsymbol{R_{\text{acc}}}$} & 
\textbf{$\boldsymbol{R_{\text{format}}}$} & 
\textbf{$\boldsymbol{R_{\text{final}}}$} & 
\textbf{Final Reward Calculation} & \textbf{Solution} \\
\midrule
$\checkmark$ & $\checkmark$ & $\checkmark$ & $\checkmark$ & $\checkmark$ & $1.00$ & $1$ & $2.00$ & $(1 + 1 + 1 + 1)/4 + 1$ & Perfect prediction \\
$\mathcal{S}$ & $\times$ & $\times$ & $\times$ & $\checkmark$ & $-0.25$ & $1$ & $0.75$ & $(-1 + 0 + 0 + 0)/4 + 1$ & Sibling error at L1 \\
$\checkmark$ & $\mathcal{S}$ & $\times$ & $\times$ & $\checkmark$ & $-0.25$ & $1$ & $0.75$ & $(1 -2 + 0 + 0)/4 + 1$ & Sibling error at L2 \\
$\checkmark$ & $\checkmark$ & $\mathcal{S}$ & $\times$ & $\checkmark$ & $-0.50$ & $1$ & $0.50$ & $(1 + 1 -4 + 0)/4 + 1$ & Sibling error at L3 \\
$\checkmark$ & $\checkmark$ & $\checkmark$ & $\mathcal{S}$ & $\checkmark$ & $-1.25$ & $1$ & $-0.25$ & $(1 + 1 + 1 -8)/4 + 1$ & Sibling error at L4 \\
\midrule
$\checkmark$ & $\checkmark$ & $\checkmark$ & $\checkmark$ & $\times$ & $1.00$ & $0$ & $1.00$ & $(1 + 1 + 1 + 1)/4 + 0$ & Correct prediction, think format error \\
$\mathcal{S}$ & $\times$ & $\times$ & $\times$ & $\times$ & $-0.25$ & $0$ & $-0.25$ & $(-1 + 0 + 0 + 0)/4 + 0$ & Sibling error at L1, think format error \\
$\checkmark$ & $\mathcal{S}$ & $\times$ & $\times$ & $\times$ & $-0.25$ & $0$ & $-0.25$ & $(1 - 2 + 0 +0)/4 + 0$ & Sibling error at L2, think format error \\
$\checkmark$ & $\checkmark$ & $\mathcal{S}$ & $\times$ & $\times$ & $-0.50$ & $0$ & $-0.50$ & $(1 + 1 -4 + 0)/4 + 0$ & Sibling error at L3, think format error \\
$\checkmark$ & $\checkmark$ & $\checkmark$ & $\mathcal{S}$ & $\times$ & $-1.25$ & $0$ & $-1.25$ & $(1 + 1 + 1 -8)/4 + 0$ & Sibling error at L4, think format error \\
- &  - & - & - & $\times$ & $0.00$ & $0$ & $0.00$ & $(0 + 0 + 0 + 0)/4 + 0$ & Invalid prediction  \\
\bottomrule
\end{tabular}
\end{table*}
\subsection{Evaluation Setup}
To systematically evaluate both in-domain classification accuracy and the model's generalization capability, we construct two evaluation sets. The base evaluation set contains samples whose category paths are fully present in the training set. In contrast, the generalization evaluation set only includes previously unseen categories to test out-of-distribution robustness.~
As illustrated in Figure~\ref{fig:evaluation-tree}, we organize categories into a hierarchical tree structure. White nodes indicate categories included during training, while gray nodes represent unseen classes that only appear in the generalization set. This setup enables us to assess how well the model can leverage structural relationships in the hierarchy to generalize to novel risk types that were not directly exposed during training.

\section{Analysis of Soft-margin Reward Value Space}\label{supp:reward}
Although the theoretical reward space under a four-level taxonomy consists of all possible combinations of category predictions across levels, the actual prediction space is significantly constrained due to our structured prompting design. Specifically, at each level, we explicitly provide the model with the definition of the parent node along with its candidate child categories, thereby restricting the model to make decisions within the correct semantic context.~
Under this hierarchical path guidance, the model almost never produces structurally inconsistent paths (e.g., predicting a parent followed by a valid child). Instead, most prediction errors are localized and involve sibling-category confusion or nearby semantic deviations. As a result, the effective reward space can be reduced to a well-defined subset of meaningful patterns.

As illustrated in Table~\ref{tab:reward-space}, the upper half of the table lists reward values where the output format is correct, ranging from fully correct paths to single-layer sibling errors. The lower half shows similar path errors but with format violations, which result in the absence of format rewards. The last row denotes completely invalid or unparseable responses. In cases where the model incorrectly classifies safe and risky content, we assign the lowest accuracy reward (\ie $R_{\text{acc}}=-1.25$) in the hierarchy to reflect the severity of such errors.

This constrained reward design not only reflects the real-world behavior of the model but also enhances the stability of GRPO training by delivering structured and informative feedback. It encourages the model to focus on fine-grained semantic distinctions between closely related categories, rather than struggling with incoherent path structures.

\end{document}